\documentclass[conference,a4paper]{IEEEtran}
\IEEEoverridecommandlockouts
\usepackage{cite}
\usepackage{amsmath,amssymb,amsfonts}
\usepackage{algorithmic}
\usepackage{graphicx}
\usepackage{textcomp}
\usepackage{xcolor}
\usepackage{hyperref}
\usepackage{lipsum}
\usepackage{multicol}
\usepackage{subcaption}
\usepackage{csquotes}
\usepackage{soul}

\usepackage[disable]{todonotes} 

\hypersetup{
    colorlinks=False,                
    breaklinks=true,
    linkcolor=blue,
    filecolor=blue,      
    urlcolor=blue,
    urlcolor = blue,
}

\def\BibTeX{{\rm B\kern-.05em{\sc i\kern-.025em b}\kern-.08em
    T\kern-.1667em\lower.7ex\hbox{E}\kern-.125emX}}
    
\begin{document}

\definecolor{black}{rgb}{0.0, 0.0, 0.0}
\definecolor{white}{rgb}{1.0, 1.0, 1.0}
\definecolor{yellow}{rgb}{1.0, 1.0, 0.8}
\definecolor{red}{rgb}{0.6, 0.0, 0.2}
\definecolor{blue}{rgb}{0.0, 0.2, 0.5}
\definecolor{green}{rgb}{0.6, 0.8, 0.8}
\definecolor{dark_green}{RGB} {0, 140, 0}
\definecolor{gold}{rgb}{0.6, 0.4, 0.1}
\definecolor{grey}{RGB}{0,0,0}
\definecolor{Gray}{gray}{0.8}
\definecolor{MediumGray}{gray}{0.9}
\definecolor{LightGray}{gray}{0.98}
\definecolor{purple}{RGB}{128,0,128}
\definecolor{sl_blue}{RGB}{47, 60, 105}
\definecolor{orange}{RGB}{255,165,0}
\definecolor{Gray}{gray}{0.85}

\newcommand{\MB}[1]{\textcolor{dark_green}{\textbf{\small [}\colorbox{yellow}{\textbf{Barni:}}{\small #1}\textbf{\small ]}}}
\newcommand{\OA}[1]{\textcolor{blue}{\textbf{\small [}\colorbox{yellow}{\textbf{Omran:}}{\small #1}\textbf{\small ]}}}
\newcommand{\TODO}[1]{\textcolor{red}{{TODO: #1}}}

\title{Detection of GAN-synthesized street videos}


\author{\IEEEauthorblockN{Omran Alamayreh\IEEEauthorrefmark{1},
Mauro Barni\IEEEauthorrefmark{2}}
\IEEEauthorblockA{Department of Information Engineering and Mathematics,
University of Siena\\
Via Roma 56, 53100 - Siena, ITALY \\
Email: \IEEEauthorrefmark{1}omran@diism.unisi.it,
\IEEEauthorrefmark{2}barni@dii.unisi.it}}

\maketitle

\begin{abstract}
Research on the detection of AI-generated videos has focused almost exclusively on face videos, usually referred to as deepfakes. Manipulations like face swapping, face re-enactment and expression manipulation have been the subject of an intense research with the development of a number of efficient tools to distinguish artificial videos from genuine ones. Much less attention has been paid to the detection of artificial non-facial videos. Yet, new tools for the generation of such kind of videos are being developed at a fast pace and will soon reach the quality level of deepfake videos. The goal of this paper is to investigate the detectability of a new kind of AI-generated videos framing driving street sequences (here referred to as DeepStreets videos), which, by their nature, can not be analysed with the same tools used for facial deepfakes. Specifically, we present a simple frame-based detector, achieving very good performance on state-of-the-art DeepStreets videos generated by the Vid2vid architecture. Noticeably, the detector retains very good performance on compressed videos, even when the compression level used during training does not match that used for the test videos.

\end{abstract}


\section{Introduction}

The recent development of Artificial Intelligence (AI) tools, enabling even non-expert users to generate fake videos of extraordinary quality, is raising increasing concerns about the credibility of digital media accessible on the internet and diffused by information and social networks. The consequences of such a loss of credibility are devastating and impact all domains of our lives, ranging from everyday life, through journalism, criminal justice, and national security. The necessity of distinguishing between fake and original media has triggered the birth of a new discipline, known as Multimedia Forensics MMF~\cite{verdoliva2020media}, which has attracted the interest of researchers and research funding agencies all around the world~\cite{SemaFor}.


With regard to the detection of fake videos generated by means of AI techniques based on Deep Learning (DL), most of the attention has been devoted to the detection of face facial videos, known as Deepfakes. As a consequence, a wide range of tools have been developed to detect several kinds of deepfake manipulations, including face swapping~\cite{Nirkin_2019_ICCV} and face re-enactment~\cite{thies2016face2face}. The interest towards this kind of videos is justified by the fact that information conveyed by speeches and facial expressions has an essential role in human social interactions. Besides, human face manipulation techniques have received an increasing attention in the last years, reaching a maturity level making them ready to be exploited in several real-life applications (including malicious ones)\cite{verdoliva2020media}. 


In comparison, the detection of non-facial artificial video sequences has received much less attention. Yet, new AI tools for the generation of non-facial videos are being developed at a fast pace and will soon reach the quality level of deepfake videos \cite{wang2018video}.
To move a first step to fill this gap, in this paper we focus on the detection of a new class of fake videos, hereafter referred to as DeepStreets, framing driving street sequences, like those described in \cite{mallya2020world}, \cite{wang2019few}, \cite{wang2018video}.

To start with, we observe that the forgery detection methods designed to work on facial videos can not be directly applied to DeepStreets. The main reason for such a difficulty is that deepfake detectors rely strongly on the facial features of the videos, all the more that they usually crop the face region before inputting the video frames to the detectors. Besides, some detectors are based on the analysis of a set of geometric and semantic features that are directly related to human faces; like eye color~\cite{matern2019exploiting} or facial pose~\cite{yang2019exposing}. No such features obviously exist in DeepStreets videos, which are characterised by a greater diversity and for which it is not possible to identify a single attention region to focus the analysis on.


To address the above limitations, we propose an end-to-end detector based on CNN. The detector works on a frame by frame basis, and does not focus on any specific region of the input frames.
To train (and test) the detector, we generated 600 videos of fake driving scenes by means of the Vid2vid architecture presented in \cite{wang2018video} and briefly summarized in Section \ref{vid_arch}. To evaluate the generalization capability of the detector and its ability to work in non-ideal conditions, we considered both raw sequences and videos compressed at different quality levels. The experiments we carried out prove the validity of the detector, which exhibited an extremely high accuracy, demonstrating that even DeepStreets kind of videos can reliably be distinguished from real videos (at least in the highly controlled setting adopted in this paper). A remarkable property of our detector is its immunity to video compression, given that the detection accuracy remains good even when the compression setting used during training is different than that used for the test videos.


The rest of this paper is structured as follows; In Section \ref{relatedwork}, we briefly review the state of the art of deepfake generation methods, video-to-video synthesis and deepfake detection methods. In Section \ref{section3}, we present the dataset we used in our experiments and the detection pipeline. The results of the experiments we carried out to validate the proposed detector are described in Section \ref{section4}. In Section \ref{section5}, we conclude the paper with some final remarks and hints for future research.

\section{related work} \label{relatedwork}


\subsection{Synthesis of deepfake videos}

With the rise of deep learning techniques, automatic generation of falsified media has reached an unprecedented level. Several manipulation approaches based on AI have been proposed in the literature for fake videos generation. These methods generally rely on two deep learning architectures; autoencoders (AE~\cite{vincent2008extracting}) and generative adversarial networks (GANs~\cite{goodfellow2014generative}). 
A popular approach to create deepfake videos is face replacement, where two autoencoders, with a shared encoder, are trained in parallel on two datasets. The datasets include face images of person \textit{A} and \textit{B}, respectively. The idea is to allow the shared encoder to learn common features for both persons \textit{A} and \textit{B} while keeping the corresponding decoders detached. To swap the faces between person \textit{A} and \textit{B}, a video framing person \textit{A} is fed to the common encoder. Then, the resulting features in the latent space are input to the decoder trained on \textit{B}. This principle is applied in several manipulation tools, like DeepFaceLab~\cite{petrov2020deepfacelab}.



In addition to autoencoders, generative adversarial networks (GANs~\cite{goodfellow2014generative}) are often used to generate falsified media. GANs consist of two networks facing each other in a zero-sum game. In the training phase, the generator network aims at producing falsified images that are undistinguishable from those contained in the training dataset of real images. Meanwhile, the discriminator network is trained to detect the images produced by the generator. After training, the generator is used to create realistic fake images that are hardly identified as such by the discriminator. GANs and autoencoders can also be used together to improve the quality of the synthesised videos, as in Faceswap-GAN~\cite{faceswap}.



\subsection{Video-to-Video synthesis} \label{vid_arch}

Most deepfake manipulation methods have been developed to create fake videos that contain human faces. Their counterpart for non-facial fake videos has been less explored. However, general-purpose video synthesis techniques are present in several forms, including video prediction~\cite{xue2016visual}, unconditional video synthesis~\cite{vondrick2016generating} and conditional video synthesis\cite{wang2018video}. The first methods produce future video frames based on past frames. The second kind of methods take in random variables and generate synthesised videos. The latter methods convert input semantic sequences into photorealistic videos. 

In our work, we used a conditional video synthesis method, known as Vid2vid~\cite{wang2018video}, to generate the training and testing datasets. Vid2vid is a video-to-video synthesis architecture operating under the generative adversarial learning framework. The goal of Vid2vid is to convert an input semantic video, for example, a sequence of semantic segmentation masks, to an output photorealistic video that represents the theme of the source video. Vid2vid is considered the counterpart of image-to-image translation methods such as Pix2pix~\cite{wang2018high} and Covst~\cite{chen2017coherent}. 
The advantage of Vid2Vid with respect to the application of image-based architectures, is that working on a frame-by-frame basis, image-based techniques do not guarantee the temporal coherence of the synthetic output video. In contrast, Vid2vid is designed in such a way to preserve the temporal dynamics of the source video.

\begin{figure}[htbp]
  \centering
  \includegraphics[scale=0.65]{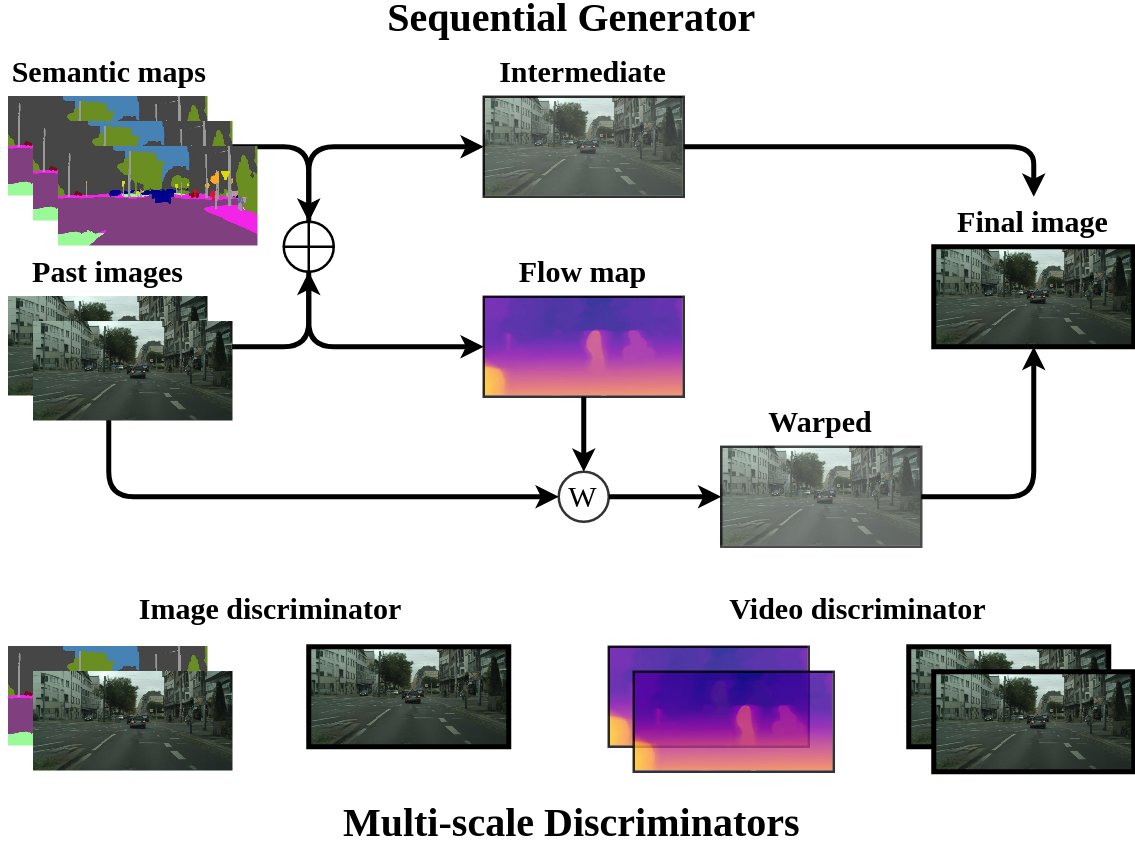}
  \caption{Vid2vid network architecture.}
  \label{vidarch}
\end{figure}

\begin{figure*}[ht]
     \centering
     \begin{subfigure}[b]{0.32925\textwidth}
         \centering
         \includegraphics[width=\textwidth]{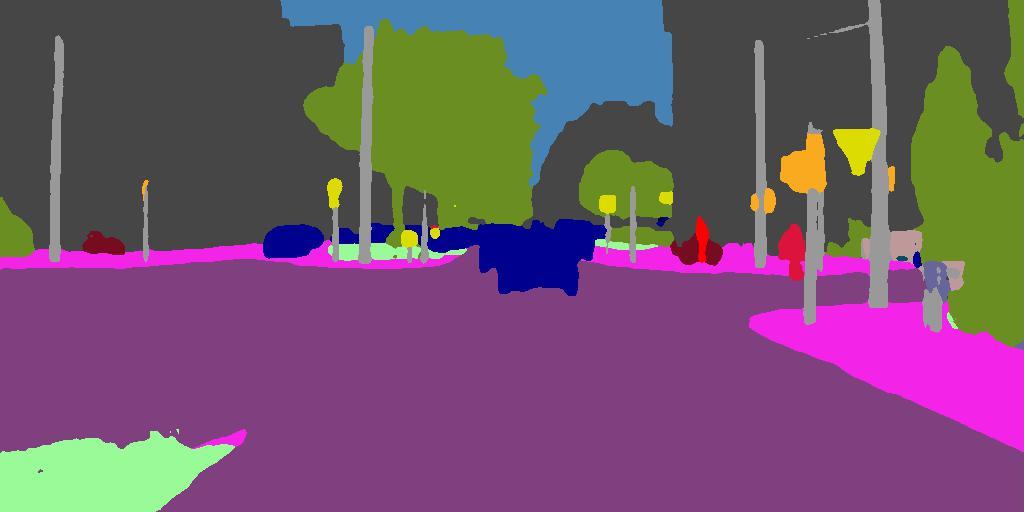}
         \caption{Semantic segmentation masks.}
         \label{fig:y equals x}
     \end{subfigure}
     \hfill
     \begin{subfigure}[b]{0.32925\textwidth}
         \centering
         \includegraphics[width=\textwidth]{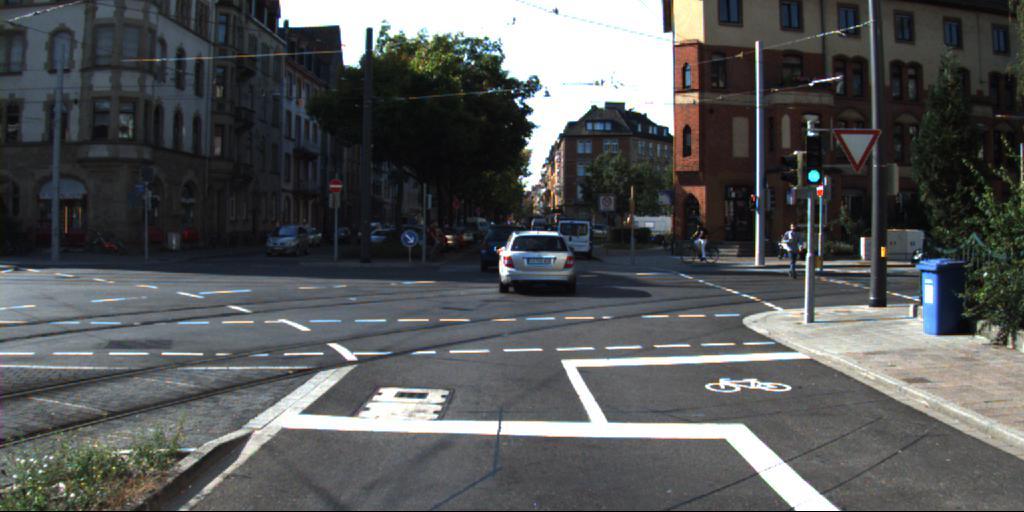}
         \caption{Sequence used to generate the mask.}
         \label{fig:three sin x}
     \end{subfigure}
     \hfill
     \begin{subfigure}[b]{0.32925\textwidth}
         \centering
         \includegraphics[width=\textwidth]{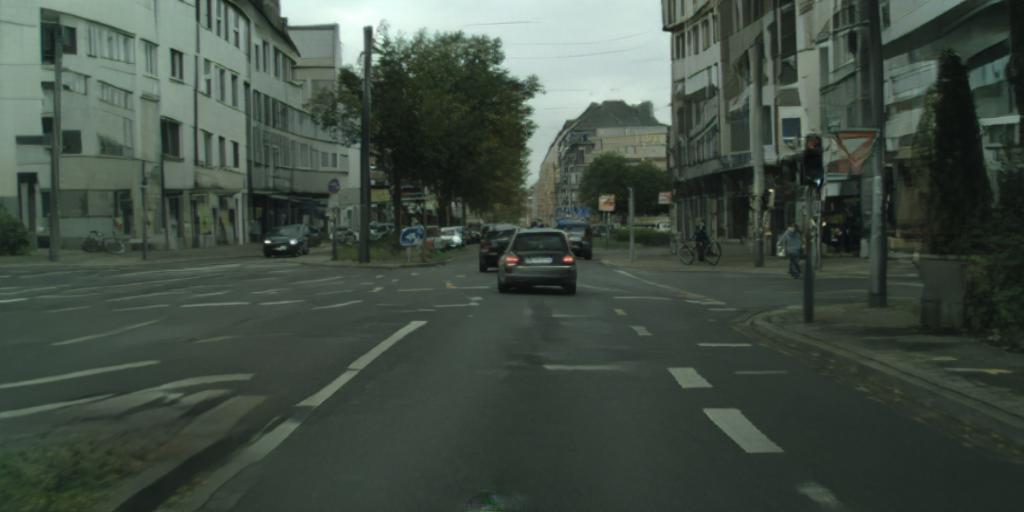}
         \caption{Generated fake video.}
         \label{fig:five over x}
     \end{subfigure}
        \caption{Example of frames generated by Vid2vid architecture.}
        \label{example_data}
\end{figure*}

The architecture of Vid2vid consists of two networks; a sequential generator and a set of multiscale discriminators. The generator takes as input a sequence of segmentation semantic maps and the frames generated previously and produces an intermediate frame and a flow map. Afterwards, the flow map is used to warp the previous frame. Then, the warped frame is combined with the intermediate frame to output the final frame, which is then used to generate the next frame and so on. The multiscale discriminators contain one discriminator for images and one for videos. The image discriminator analyses the input semantic map and the output images to ensure that each output frame resembles a real image. Simultaneously, the video discriminator takes in the flow maps and neighbouring frames to guarantee the temporal coherence of the generated videos. The architecture of Vid2vid is depicted in figure \ref{vidarch}.

Recently, two versions of Vid2vid have been released: 
Few-shot video-to-video synthesis~\cite{wang2019few} and World-Consistent Video-to-Video Synthesis~\cite{mallya2020world}. The first work increases the generalisation capability of Vid2vid by using a novel network weight generation module based on attention mechanism. The latter improves the generated videos' temporal consistency by exploiting, so-called, \textit{guidance images}, which are a consolidation of the \textit{3D world} rendered in the previous frames. In our experiments, we used Vid2vid and Wc-vid2vid architectures to generate DeepStreets dataset, as described in section \ref{section3}.

\subsection{Deepfake detection methods}

The techniques developed so far to distinguish real and deepfake videos belong to two main classes: those based on handcrafted features and those relying on end-to-end completely automatic detectors based on DL. On the one hand, handcrafted methods exploit a set of predefined traces that result from the production pipeline of deepfakes, like, visual artifacts~\cite{li2018exposing}, semantic inconsistencies~\cite{matern2019exploiting}.
On the other hand, DL-based methods, specifically those based on convolutional neural networks (CNN), deliver end-to-end solution to the forgery detection problem. Approaches like MesoInception~\cite{afchar2018mesonet}, and XceptionNet~\cite{rossler2019faceforensics++}, provide superior performance compared to the traditional handcrafted methods~\cite{rossler2019faceforensics++}. CNN based methods still suffer from significant shortcomings like the difficulty of interpreting the output they provide, the vulnerability to adversarial attacks, and the difficulty to maintain good performance when the training and test phases work in mismatched conditions. Moreover, the performance of DL-based methods drop in the presence of video compression~\cite{verdoliva2020media}. 

Here we propose a forgery detector that relies on CNN. As demonstrated in section~\ref{section4}, the detector shows impressive performance, and a good degree of robustness against video coding, while its generalization capability strongly depends on the experimental conditions.

\section{methodology} \label{section3}

\subsection{DeepStreets Dataset} \label{deepdataset}

To generate the dataset we used in our experiments, we started from the pre-trained models of Vid2vid and Wc-vid2vid on the Cityscapes dataset \cite{cordts2016cityscapes}. Both models require a sequence of semantic segmentation masks as input to generate a video. Additionally, Wc-vid2vid requires a sequence of guidance images to improve the output video's temporal consistency~\cite{mallya2020world}. The models’ output is a photorealistic video of street scenes that render the content of the input sequence of the semantic segmentation masks. Figure \ref{example_data} shows an example of the generated videos.

We used Cityscapes~\cite{cordts2016cityscapes} and Kitti~\cite{Geiger2013IJRR} datasets to produce the input segmentation maps. The cityscapes dataset consists of diverse urban street videos from several German cities. The videos have $2048\times 1024$ resolution captured using a pair of stereo cameras at varying times of the year. The Kitti dataset contains driving videos that have been captured near Karlsruhe, Germany, using colour and grayscale camera images. In our experiments, we used only the colour videos with $1392\times 512$ resolution.

Since not all the images of Cityscapes are labelled with segmentation masks, and to be consistent with Vid2vid pipeline, we have annotated the images of Cityscapes and Kitti by using the network from Zhuet \textit{et al.}~\cite{zhu2019improving}. We also followed Wc-vid2vid pipeline to generate the guidance images, and we performed \textit{structure from motion} SfM on the video sequences using OpenSfM~\cite{sfm}.

We divided the dataset into three subsets; cityvid, Citywcvid and Kittivid. We generated Cityvid by inputting the segmentation masks of Cityscapes into Vid2vid, and Citywcvid by feeding Wc-vid2vid with segmentation masks and guidance images of Cityscapes. For Kittivid, we used the segmentation masks of Kitti dataset as input to Vid2vid. Each sub-dataset contains 400 videos, half of them are fake, and the other half are real (the real videos are the original sequences from Cityscapes and Kitti datasets). The videos resolution is $512\times 1024$ and their duration is 3 seconds, with 30 frames. Table \ref{table:1} summarises the characteristics of the sub-datasets. 


\begin{table}[htbp]

\setlength{\tabcolsep}{10pt}
\renewcommand{\arraystretch}{1.5}
\centering
\begin{tabular}{l|ll}
\hline 
\textbf{sub-dataset} & network & segmentation map\\ 
\hline
Cityvid & vid2vid & cityscapes\\

Citywcvid & wc-vid2vid & cityscapes\\ 
Kittivid & vid2vid & kitti\\ 
\hline 
\end{tabular}

\captionsetup{justification=centering}
\caption{A summary of DeepStreets dataset.}
\label{table:1}
\end{table}

In addition to the raw videos produced by the Vid2vid models, we have compressed the videos with two quality levels; \textit{HQ}, and \textit{LQ}. In particular, we compressed the raw videos by using the H.264 codec with a constant rate quantisation parameter equal to 23 and 40, respectively. The compressed videos as well as the raw videos are available on our website; \url{http://clem.dii.unisi.it/~vipp/datasets.html}.


\subsection{The Proposed Detector} \label{tpd}

To distinguish genuine videos from DeepStreets ones, we propose to use a data-driven forgery detector based on CNN. Since the proposed dataset has no clear predefined traces, it is difficult to rely on handcrafted features to differentiate between fake and genuine videos. In addition, the existing deepfake detectors usually detect and crop the faces in each frame of the videos. In our case, we need to input the whole frame to the detector, thus making a CNN-like architecture a good candidate for our task. Having said that, we used Xception network~\cite{chollet2017xception} as the backbone for our detector. 
 
Xception is a CNN-like architecture based on depthwise separable convolution layers. It uses 36 convolutional layers for the feature extraction part. The layers are organised into 14 modules; each module has linear residual connections, excluding the first and last modules. The convolutional part is followed by an optional fully-connected layer and a logistic regression layer. For our detector, we added a fully-connected layer with two outputs on top of the feature extraction part.

\begin{figure}[htbp]
  \centering
  \includegraphics[scale = 0.49]{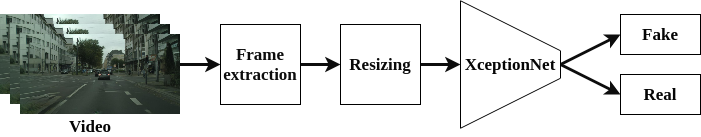}
  \caption{Detection pipeline.}
  \label{pipline}
\end{figure}

We formalise the detection task as a \textit{per-frame binary classification} problem. We start by extracting the frames from each video. Then, to limit the computational burden, we resized each frame to $256\times 512$ resolution. Afterwards, the frames are sequentially fed to the Xception network. The detection pipeline is shown in figure \ref{pipline}.

To ensure the reproducibility of our results, we report the setting that we have used to conduct the experiments. We split the dataset into training, testing, and validation sets, with ratios of 60\%, 25\%, and 15\%, respectively. All the results reported hereafter regard the detection accuracy on the test set. During the training, we used Adam optimizer with learning rate $(\alpha)$ = 0.0001 and the default values for the first and second-order moments $\left(\beta_{1}=0.9, \beta_{2}=0.999\right)$, and an epsilon value $(\epsilon)$ = $10^{-8}$. Finally, the batch size was set to 8. Training was stopped whenever validation loss did not decrease for 10 consecutive epochs. 
All the experiments were performed using PyTorch framework on a workstation with one intel core i9 and four NVIDIA GeForce RTX 2080 Ti GPUs.


\section{Experimental results} \label{section4}

We have conducted several experiments to assess the performance of the proposed forgery detector on DeepStreets dataset at various compression levels, including matched and mismatched conditions. Furthermore, we have investigated its generalization capability by carrying out cross-dataset analysis between all subsets of DeepStreets dataset.

\begin{table}[htbp]

\setlength{\tabcolsep}{10pt}
\renewcommand{\arraystretch}{1.5}

\centering
\begin{tabular}{l|lll}

\hline 
\textbf{Dataset\textbackslash Quality} \rule{3ex}{0pt} & RAW \rule{4ex}{0pt} & HQ \rule{4ex}{0pt} & LQ \\
\hline
Cityvid & 100.00 & 100.00 & 97.93 \\
Citywcvid & 99.76 & 99.62 & 99.96 \\ 
Kittivid & 100.00 & 100.00 & 100.00 \\ 
DeepStreets & 99.86 & 100.00 & 99.19 \\ 
\hline 
\end{tabular}

\captionsetup{justification=justified,singlelinecheck=false}
\caption{Detection accuracy (the detector is asked to distinguish the original sequences used to create the segmentation masks and the synthetic videos).}
\label{table:2}
\end{table}

In our first experiments, we trained the detector to distinguish the original sequences used to generate the segmentation masks and the sequences produced by the synthetic video generators. Table \ref{table:2} shows the detection accuracy of the detector on all subsets of DeepStreets dataset. The results refer to matched training and testing, that is when training and testing is carried out on the same subsets. As it can be seen, the classification accuracy is almost perfect for all the subsets even when training and testing is carried out on the whole dataset. With regard to the impact of video compression on detection accuracy, we carried out two sets of experiments. In Table \ref{table:2}, the detector is trained and tested in matched compression conditions, while in Table \ref{table:3} different training and testing conditions are used. In both cases, compression has a minor impact on the accuracy. This marks a difference with respect to deepfakes videos, where the presence of compression leads to a significant performance drop (see for instance the results reported in \cite{rossler2019faceforensics++} with regard to low-quality videos).

\begin{table}[htbp]

\setlength{\tabcolsep}{10pt}
\renewcommand{\arraystretch}{1.5}

\centering
\begin{tabular}{l|lll}
\hline 
\textbf{Training\textbackslash Testing } \rule{2ex}{0pt} & RAW \rule{4ex}{0pt} & HQ\rule{4ex}{0pt} & LQ \\ 
\hline
RAW & 99.89 & 99.90 & 95.41 \\
HQ & 100.00 & 100.00 & 95.72 \\ 
LQ & 99.70 & 99.63 & 99.19 \\ 
\hline 
\end{tabular}

\captionsetup{justification=justified,singlelinecheck=false}
\caption{Compression mismatch experiments. The detector has been trained (and tested) on the entire DeepStreet dataset.}
\label{table:3}
\end{table}

To investigate the generalisation capability of the proposed forgery detector, we performed a cross-dataset analysis between all the subsets of DeepStreets dataset. As illustrated in Table \ref{table:4}, in some cases, dataset mismatch causes a dramatic drop of the performance, while in other cases the performance loss is less significant. Such behaviour can be explain by the different production pipeline used for the three sub-datasets. As mentioned in section~\ref{deepdataset}, we have generated each subset by adopting different settings, including the source of the semantic maps and the specific Vid2vid architecture that generated the videos. 

\begin{table}[htbp]

\setlength{\tabcolsep}{10pt}
\renewcommand{\arraystretch}{1.5}

\centering
\begin{tabular}{l|lll}
\hline 
\textbf{Training\textbackslash Testing} \rule{2ex}{0pt} & Cityvid \rule{3ex}{0pt} & Citywcvid\rule{3ex}{0pt} & Kittivid \\ 
\hline
Cityvid & 100.00 & 71.50 & 88.16 \\
Citywcvid & 98.76 & 99.76 & 50.00 \\ 
Kittivid & 50.03 & 50.00 & 100.00 \\ 
\hline 
\end{tabular}

\captionsetup{justification=justified,singlelinecheck=false}
\caption{Cross-dataset analysis between subsets of DeepStreets dataset (RAW videos).}
\label{table:4}
\end{table}

For instance, we can see that performance significantly deteriorates if we perform cross-dataset analysis between Kittivid and Citywcvid sub-datasets, since these  sub-datasets have a different source for the semantic sequences. We used Kitti~\cite{Geiger2013IJRR} and Cityscapes~\cite{cordts2016cityscapes} dataset to generate Kittivid and Citywcvid sub-datasets, respectively. Besides, the Vid2vid architectures used to generate them are different (see table~\ref{table:1}). In contrast, the performance is quite adequate when we perform cross-dataset analysis between Cityvid and Citywcvid sub-datasets since both of them have the same source of semantic sequences (Cityscapes~\cite{cordts2016cityscapes} in this case) and differ only for the Vid2vid architecture used for the generation.

We also conducted a second set of experiments, wherein the detector is asked to distinguish the synthetic sequences from the real sequences used to train the generators (namely the sequences from real images of the Cityvid sub-dataset). The reason why we carried out this additional experiment is that the generators tend to produce scenes with colours and light conditions similar to those of the sequences used to train them (that is Cityvid real images), so it may be relatively easy for the discriminator to distinguish original images belonging to a different dataset, e.g. Kitti sequences, since they are characterised by different environmental conditions (light and sun conditions, for instance). This is not the case, when the real images are taken from the same dataset used to train the GANs.
In the new setting, the real videos are driving scenes from the Cityscapes dataset, and the fake videos are synthesised videos generated by using the semantic sequences from the Kitti dataset. The detector trained in this way achieved 100\% accuracy on the test set of the new sub-dataset.
We also tested the ability of the detector to distinguish between real Citivyd videos and fake videos generated starting from segmentation maps of Citivyd (the Cityvid sub-datasets in Table \ref{table:1}).
Interestingly, the detector achieved a 80.80\% accuracy, showing a certain capability to detect examples of fake videos that are not represented in the training set.


\section{Conclusions} \label{section5}

We have demonstrated the possibility of detecting non-facial fake videos generated by video-to-video translations tools by using data-driven methods, even when training and testing are carried out in compression mismatched conditions. 
We chose the street views as an example of non-facial fake videos. Nevertheless, the generation and detection pipeline can be extended to other contents. Future research will be devoted to create a dataset containing street videos and other non-facial fake videos, such as F1 videos, tennis videos etc. (despite the computational burden required to train new video-to-video models \cite{wang2018video}). Another direction for future research is to investigate the generalization capability of the proposed forgery detector to a class of facial and non-facial synthetic videos that is not included in the training set. Finally, it would also be interesting to asset the robustness of the proposed detector against adversarial attacks.

\section*{Acknowledgment}

This work has been supported by the PREMIER project under contract PRIN 2017 2017Z595XS-001, funded by the Italian Ministry of University and Research and by the Defense Advanced Research Projects Agency (DARPA) and the Air Force Research Laboratory (AFRL) under agreement number FA8750-20-2-1004. The U.S. Government is authorized to reproduce and distribute reprints for Governmental purposes notwithstanding any copyright notation thereon. The views and conclusions contained herein are those of the authors and should not be interpreted as necessarily representing the official policies or endorsements, either expressed or implied, of DARPA and AFRL or the U.S. Government.



\bibliographystyle{unsrt}
\bibliography{Bibliography}


\end{document}